%% file: arxiv.tex
\documentclass[conference]{IEEEtran}
\IEEEoverridecommandlockouts

\usepackage{amsmath,amssymb,amsfonts}
\usepackage{graphicx}
\usepackage{booktabs}
\usepackage{tikz}
\usetikzlibrary{positioning,arrows.meta,fit,backgrounds}
\usepackage{pgfplots}
\pgfplotsset{compat=1.16}
\usepackage{url}
\usetikzlibrary{calc} 
\newcommand{\microF}{micro-F1}
\newcommand{\macroF}{macro-F1}
\usepackage{pifont}
\newcommand{\cmark}{\ding{51}}
\newcommand{\xmark}{\ding{55}}
\usepackage{listings}
\usepackage{xcolor}
\usepackage{booktabs}
\usepackage{hyperref}
\usepackage{float}

\begin{document}

\title{Graph-Constrained Policy Learning for Extreme Clinical Code Prediction}

\author{\IEEEauthorblockN{Amritpal Singh, Sebastian Torres, Khawar Shakeel, Syed Ahmad Chan Bukhari}
\IEEEauthorblockA{Division of Computer Science, Mathematics and Science\\
St. John's University, New York, NY, USA\\
bukharis@stjohns.edu}}

\maketitle

\begin{abstract}
Clinical code prediction maps unstructured discharge summaries to ICD-10-CM leaf codes from a large, sparse, and deeply hierarchical label space. Most automated coding systems treat this problem as flat multi-label classification, scoring each code independently and leaving rare labels with limited training signal. We propose a graph-constrained traversal policy that reformulates ICD prediction as a finite-horizon decision process over a pruned code hierarchy. A single language model descends the graph level by level, selecting valid child nodes from the current frontier until billable leaf codes are reached. This constraint converts extreme multi-label prediction into a sequence of sparse, hierarchy-aware subset decisions while guaranteeing structurally valid outputs.

On MIMIC-IV discharge summaries, our best supervised traversal policy, SFT-1$^{+}$, achieves 0.709 micro-F1 on a curated 50-code subset and 0.527 micro-F1 on the full 15{,}761-code label space, outperforming flat baselines including CAML, LAAT, and PLM-ICD. The gains are largest in the full setting, where SFT-1$^{+}$ improves over the strongest flat baseline by $+0.044$ micro-F1 and $+0.157$ macro-F1, suggesting that graph-constrained decomposition mitigates the rare-code bottleneck. We further conduct a controlled factorial study over architecture, training algorithm, and data budget. Across both scales, a single shared policy matches a three-specialist cascade while avoiding its context-window overflow failure on 28--32\% of full-space test notes; increasing supervised trajectory data is the only intervention that consistently improves performance; and GRPO reinforcement learning provides no benefit over supervised continuation under matched data. These results show that simple graph-constrained policy learning can outperform more complex flat, cascaded, and reinforcement-learning alternatives for extreme clinical code prediction.
\end{abstract}

\section{Introduction}

Clinical code prediction maps unstructured documentation to standardized diagnosis codes used for reimbursement, quality measurement, cohort discovery, epidemiology, and risk adjustment. In U.S. inpatient care, diagnoses are recorded using ICD-10-CM, a deeply hierarchical system that refines broad chapters into sections, subsections, categories, subcategories, and billable leaf codes.

Automated ICD coding from discharge summaries is challenging because it combines long-document understanding with extreme multi-label prediction. Notes may contain several thousand tokens, while the label space is large, sparse, and heavily imbalanced: common conditions appear frequently, but many clinically important leaf codes occur rarely. Prior work has explored linear classifiers, CNNs, RNNs, label-wise attention, transformer encoders, graph-enhanced architectures, and retrieval-augmented systems~\cite{caml,multirescnn,laat,hypercore,xmltransformer}; however, most methods still treat coding as flat multi-label classification over leaf codes.

This paper casts ICD-10-CM code prediction as graph-constrained traversal, formalized as a finite-horizon MDP over the pruned ICD-10-CM hierarchy (Section~\ref{sec:mdp}). A single decoder-only language model descends the hierarchy level by level. At each depth, a lightweight environment exposes the valid children of the current frontier through tool calls; the model selects the children supported by the note, and the descent continues until billable leaves are reached. Hard validity constraints restrict each selection to the candidate set $C_t$, so structurally invalid codes receive zero probability by construction. Instead of scoring the full label space, $O(|L|)$, for each note, the traversal policy reduces inference to $O(\sum_{\ell}|C_\ell|)$, where $|C_\ell| \ll |L|$ after pruning to relevant branches. Extreme multi-label classification therefore becomes a short sequence of sparse, graph-constrained subset decisions. As Table~\ref{tab:positioning} details, our system combines hierarchy-guided inference, hard structural constraints, a trained policy, multi-turn context across depths, and formal MDP grounding; prior systems omit at least one of these properties.

On MIMIC-IV discharge summaries, the trained traversal policy significantly outperforms three strong flat baselines, CAML~\cite{caml}, LAAT~\cite{laat}, and PLM-ICD~\cite{plmicd}, on both a curated 50-code subset and the full 15{,}761-code ICD-10-CM space. On the full label space, SFT-1$^{+}$ exceeds the strongest flat baseline, LAAT, by $+0.044$ \microF{} and $+0.157$ \macroF{}. CAML and PLM-ICD collapse to 0.002 \macroF{}, indicating limited ability to discriminate among 15{,}761 candidates without structural guidance. These results show that graph-constrained decomposition partially insulates against the rare-code bottleneck.

We further ask whether additional architectural or algorithmic complexity is warranted once graph-constrained traversal is introduced. We conduct a controlled factorial study on the same MIMIC-IV data, isolating architecture, training algorithm, and annotation budget: one shared traversal policy versus a three-specialist cascade, supervised fine-tuning versus GRPO reinforcement learning, and 2{,}000 versus 5{,}000 gold trajectories. Three findings hold at both label-space scales. First, the shared policy matches the cascade on accuracy while avoiding the cascade's context-window overflow failure on 28--32\% of full-space test notes. Second, additional supervised trajectories are the only intervention that reliably improves performance, with gains roughly doubling from the 50-code to the full-space setting. Third, GRPO does not improve over supervised continuation under matched data. Together, these findings support a simple practical conclusion: use one shared traversal policy and invest in supervised trajectories rather than cascaded architectures or reinforcement learning.

\section{Related Work}

\subsection{Hierarchical and Graph-Aware Coding}
Because ICD is hierarchical, several models incorporate code structure. HyperCore represents ICD codes with hyperbolic and co-graph embeddings to exploit hierarchical relations and co-occurrence patterns~\cite{hypercore}. Later graph-based and knowledge-guided models~\cite{msmn,ehrgraph,knowledgeprompt} use code descriptions, ancestor relations, and label graphs to improve rare-code generalization. These methods show that ICD structure carries predictive signal, but most use the hierarchy as an auxiliary representation or regularizer rather than traversing it directly. In Section~\ref{sec:results}, we show that direct graph-constrained traversal yields larger gains as the label space grows.

\subsection{Clinical Foundation Models and LLM Coding}
Clinical language models such as ClinicalBERT~\cite{clinicalbert} and BioBERT~\cite{biobert} showed that biomedical pretraining improves representation learning for healthcare NLP. For ICD coding, transformer encoders~\cite{plmicd} and generative prompting have been proposed as alternatives to CNN/RNN pipelines, but long discharge notes and large code vocabularies remain limiting factors. Prompt-only LLM coding also raises reliability concerns~\cite{soroush}, including nonexistent codes, mixed ICD revisions, or plausible but invalid descendants. Closest to our setting, Boyle et al.~\cite{boyle} use the ICD hierarchy to search sparsely rather than score the full vocabulary, but rely on prompting alone. We instead bind generation to an explicit candidate set from the current ICD-10-CM graph and train the traversal policy on oracle demonstrations.

\subsection{Positioning}
\label{sec:positioning}
Table~\ref{tab:positioning} situates our approach against representative prior systems along five axes that together define the design space for hierarchical ICD coding.

\begin{table}[t]
\caption{Design-space comparison of representative ICD coding systems. \cmark{} = yes, \xmark{} = no, \textasciitilde{} = partial.}
\label{tab:positioning}
\centering
\setlength{\tabcolsep}{3pt}
\scriptsize

\begin{tabular}{lccccc}
\toprule
& \rotatebox{65}{Hierarchy at inference}
& \rotatebox{65}{Hard constraint}
& \rotatebox{65}{Trained policy}
& \rotatebox{65}{Multi-turn context}
& \rotatebox{65}{Formal MDP} \\
\midrule
CAML~\cite{caml}                 & \xmark              & \xmark & \cmark & \xmark & \xmark \\
LAAT~\cite{laat}                 & \xmark              & \xmark & \cmark & \xmark & \xmark \\
PLM-ICD~\cite{plmicd}           & \xmark              & \xmark & \cmark & \xmark & \xmark \\
HyperCore~\cite{hypercore}       & \textasciitilde{}   & \xmark & \cmark & \xmark & \xmark \\
MSMN~\cite{msmn}                 & \textasciitilde{}   & \xmark & \cmark & \xmark & \xmark \\
HiCu~\cite{hicu}                 & \cmark              & \xmark & \cmark & \xmark & \xmark \\
Boyle et al.~\cite{boyle}        & \cmark              & \cmark & \xmark & \cmark & \xmark \\
Ours                             & \cmark              & \cmark & \cmark & \cmark & \cmark \\
\bottomrule
\end{tabular}

\end{table}


Our formulation combines all five properties: the ICD-10-CM graph is traversed at inference time with hard validity constraints, the traversal policy is trained directly on oracle demonstrations, decisions at each depth condition on the full accumulated history, and the procedure is grounded in a formal MDP (Section~\ref{sec:mdp}).

\section{Data and Task Construction}

\subsection{Source and Cohort Construction}
Following the MIMIC-IV coding setup of Boyle et al.~\cite{boyle}, we construct our dataset from MIMIC-IV~\cite{mimic4} and MIMIC-IV-Note~\cite{mimic4note}, joining admissions, diagnosis codes, and discharge notes by admission identifier. We retain admissions with non-null discharge notes coded exclusively under ICD-10, discarding any admission containing at least one ICD-9 code to avoid mixing code schemas. This yields 122{,}281 admissions, each represented as a discharge note and gold diagnosis code set.

We then validate diagnosis codes against the official active ICD-10-CM code list (April 2026 release, 74{,}719 codes), retaining only admissions whose gold codes are present in this reference set. This removes admissions with deprecated, malformed, or non-billable codes, leaving 122{,}197 notes. This validated cohort serves as the base population for both label spaces and downstream splits. All data access, processing, and model training were conducted within an AWS SageMaker instance operating under a Business Associate Agreement (BAA) with Amazon, in compliance with MIMIC-IV data use requirements.

\subsection{Label Space and Code Graph}
Similar to Boyle et al.~\cite{boyle}, we define two label spaces over the validated cohort: \emph{all}, containing every observed code with 15{,}761 leaves, and \emph{frequent}, containing the 50 most common codes. These provide controlled settings at different points on the difficulty--realism trade-off. For each setting, we build a pruned ICD-10-CM graph retaining gold codes and their ancestors up to the chapter level, while discarding unreferenced branches (Table~\ref{tab:splits}).

\begin{table}[t]
\caption{Label space, code graph, and split statistics.}
\label{tab:splits}
\centering
\setlength{\tabcolsep}{3pt}
\scriptsize
\begin{tabular}{lccccccc}
\toprule
Variant & Leaves & Nodes & Ch. & SFT & Cont. & Test & Avg.\ gold/note \\
\midrule
All      & 15{,}761 & 26{,}413 & 21 & 2{,}000 & 3{,}000 & 1{,}000 & 13.2 \\
Frequent & 50       & 146      & 11 & 2{,}000 & 3{,}000 & 1{,}000 & 5.3 \\
\bottomrule
\end{tabular}
\end{table}

\subsection{Splits and Statistics}
\label{sec:splits}
For each label space, we sample 6{,}500 notes from the validated cohort and partition them into four disjoint splits: 2{,}000 for supervised fine-tuning, 3{,}000 for continuation training, 500 for validation, and 1{,}000 for testing. The continuation split is used for both additional supervised fine-tuning and GRPO reinforcement learning, supporting the factorial design in Section~\ref{sec:arms}. Splitting is performed at the note level with seed 42 so that no admission appears in more than one split. Test and validation sets are held fixed across both label spaces and all training arms.

Table~\ref{tab:splits} reports summary statistics. Notes in the \emph{all} setting carry more gold codes on average than in the \emph{frequent} setting because \emph{frequent} includes only diagnoses falling within the top-50 codes. Because ICD coding comparisons are sensitive to cohort construction, split strategy, and code-version handling~\cite{edin,mimic4icd}, we fix note-level splits, evaluate both label spaces, and hold test and validation sets constant across all arms.

\subsection{Gold Trajectory Construction}
\label{sec:gold}
Each training example is a multi-turn gold trajectory demonstrating the correct graph traversal for a note. We expand each gold ICD-10-CM code to its full ancestor path, from chapter root to billable leaf, and replay the resulting paths through the traversal environment. At each turn, the environment presents candidate nodes and the oracle selects exactly those on a gold ancestor path. Leaf codes are recorded automatically upon expansion, and the trajectory ends with an \texttt{ANSWER} turn listing all recorded leaf codes.

This yields one deterministic sequence per note, with assistant tool calls and environment responses tracing every gold code from root to leaf. During tokenization, environment-response tokens are masked from the training loss, so the model is supervised only on its own decisions: which nodes to select and when to answer.

\section{Methodology}

\subsection{Traversal as a Markov Decision Process}
\label{sec:mdp}

We formalize ICD-10-CM code prediction as a finite-horizon MDP over the pruned ICD-10-CM graph.
$\mathcal{G}=(\mathcal{V},\mathcal{E})$,
where $\mathcal{V}$ is the set of code-graph nodes and $\mathcal{E}$ the directed parent$\rightarrow$child edges.

\textbf{State.}
A state $s_t=(x,F_t,R_t)$ comprises the discharge note $x$ (fixed throughout the episode), the \emph{frontier}
$F_t\subseteq\mathcal{V}$ of nodes pending expansion, and the set $R_t$ of billable leaf codes recorded so far.
The initial state is
$s_0=(x,\{\textsc{root}\},\emptyset)$.

\textbf{Action.}
At each step the policy first expands every node in $F_t$, revealing the candidate set
\[
C_t=\bigcup_{v\in F_t}\mathrm{children}_{\mathcal{G}}(v).
\]
The action $a_t\subseteq C_t$ is the subset of children the note supports.
The constraint $a_t\subseteq C_t$ is enforced by the environment, so structurally invalid selections receive zero probability by construction rather than by penalty.

\textbf{Transition.}
The transition is deterministic.
Any selected node that is a billable leaf is added to $R_t$; the remainder form the next frontier:
\begin{align*}
F_{t+1} &= \{v\in a_t\mid v\text{ is not a leaf}\},\\
R_{t+1} &= R_t\cup\{v\in a_t\mid v\text{ is a leaf}\}.
\end{align*}
The episode terminates when $F_{t+1}=\emptyset$.

\textbf{Reward.}
For supervised fine-tuning the reward is implicit: the policy is trained to imitate oracle trajectories constructed from the gold code sets (Section~\ref{sec:gold}).
For reinforcement learning (GRPO-1, Section~\ref{sec:arms}), we define an explicit terminal reward
\[
r(s_T)=\tfrac12\left[\mathrm{set\mbox{-}F1}(R_T,C^*)
+\mathrm{path\mbox{-}F1}(R_T,C^*)\right],
\]
where $C^*$ is the gold code set and path-F1 awards partial credit for correct ancestor paths (Equation~\ref{eq:path}).
Rewards are zero for all non-terminal transitions; only the terminal state receives the reward defined above.

\textbf{Policy.}
The policy $\pi_\theta(a_t\mid s_t)$ is parameterized by a single QLoRA adapter over a decoder-only backbone.
Because the full traversal history
$(a_0,C_0,\ldots,a_{t-1},C_{t-1})$
is retained in the context window, the Markov property holds with respect to the model's input: each action conditions on the note, the complete descent so far, and the current candidate set.

The horizon is bounded by the depth of the deepest selected path in $\mathcal{G}$ (typically 5--7 expand-and-select rounds).
The effective action space at each step satisfies
$|C_t|\ll|\mathcal{V}|$,
reducing the combinatorial multi-label problem to a short sequence of sparse, graph-constrained subset selections.
Section~\ref{sec:env} describes the concrete implementation of this MDP as a tool-calling environment.

\subsection{Traversal Environment}
\label{sec:env}
We implement the MDP defined in Section~\ref{sec:mdp} as a guided descent through the  pruned ICD-10-CM graph, mediated by a lightweight environment that maintains a traversal agenda and exposes two tools and a terminal action:
\begin{itemize}
\item \texttt{expand\_level()} pops all currently pending nodes from the agenda and reveals their children, grouped by parent. If a pending node is a billable leaf (a code with no descendants in the pruned ICD-10-CM graph) it is recorded automatically as a final answer rather than expanded, since there is nothing below it to drill into. When the agenda is empty, the environment returns \texttt{DONE} and lists all recorded codes.
\item \texttt{select(children=[...])} accepts a list of child IDs drawn exclusively from the most recent \texttt{expand\_level()} output and queues them as pending for the next expansion. The environment validates every ID against the set of children just presented; invalid IDs are rejected, which realizes the constrained-decoding principle of restricting generation to a valid candidate set~\cite{guided} at the level of tool arguments. An empty list is permitted, allowing the model to skip an entire level when no candidates are supported by the note.
\item \texttt{ANSWER: code1, code2, ...} terminates the trajectory and emits all recorded leaf codes as the model's final prediction.
\end{itemize}

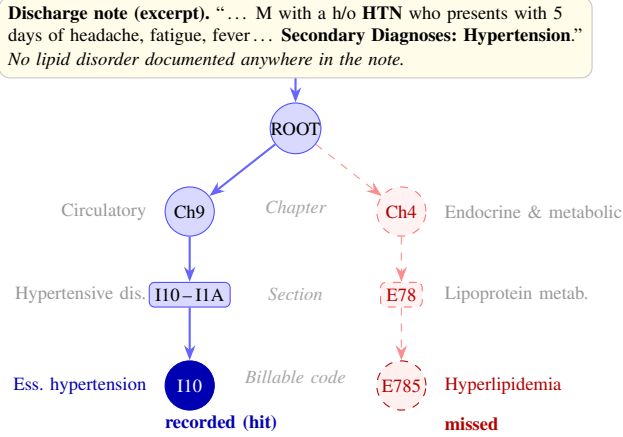
\begin{figure}[t]
  \centering
  \begin{tikzpicture}[
    every node/.style={font=\scriptsize},
    selc/.style={circle,draw=blue!70,fill=blue!15,minimum size=6.5mm,inner sep=0pt},
    missc/.style={circle,draw=red!55,dashed,fill=red!5,text=red!70!black,minimum size=6.5mm,inner sep=0pt},
    selr/.style={rounded corners=2pt,draw=blue!70,fill=blue!15,inner sep=2.2pt},
    missr/.style={rounded corners=2pt,draw=red!55,dashed,fill=red!5,text=red!70!black,inner sep=2.2pt},
    leaf/.style={circle,draw=blue!70!black,fill=blue!70!black,text=white,minimum size=6.5mm,inner sep=0pt},
    miss/.style={circle,draw=red!70!black,dashed,fill=red!6,text=red!70!black,minimum size=6.5mm,inner sep=0pt},
    note/.style={rectangle,draw=gray!45,fill=yellow!10,rounded corners,align=left,
                 text width=76mm,inner sep=3.5pt,font=\scriptsize},
    dL/.style={font=\scriptsize,text=gray!80,anchor=east},
    dR/.style={font=\scriptsize,text=gray!80,anchor=west},
    lvl/.style={font=\scriptsize\itshape,text=gray!70,anchor=east},
    edge/.style={-{Stealth[length=2mm]},thick,blue!60},
    medge/.style={-{Stealth[length=2mm]},red!45,dashed},
  ]
  \node[note] (note) at (3.0,3.75)
    {\textbf{Discharge note (excerpt).} ``\,\ldots\ M with a h/o \textbf{HTN} who presents
     with 5 days of headache, fatigue, fever\,\ldots\ \textbf{Secondary Diagnoses: Hypertension}.''
     \\[1pt]\emph{No lipid disorder documented anywhere in the note.}};
  \node[selc] (root) at (3.0,2.55) {ROOT};
  \node[selc]  (ch9) at (1.6,1.45) {Ch9};
  \node[missc] (ch4) at (4.4,1.45) {Ch4};
  \node[selr]  (i1x) at (1.6,0.35) {I10\,--\,I1A};
  \node[missr] (e78) at (4.4,0.35) {E78};
  \node[leaf] (i10)  at (1.6,-0.85) {I10};
  \node[miss] (e785) at (4.4,-0.85) {E785};
  \draw[edge]  (note) -- (root);
  \draw[edge]  (root) -- (ch9);
  \draw[medge] (root) -- (ch4);
  \draw[edge]  (ch9) -- (i1x);
  \draw[medge] (ch4) -- (e78);
  \draw[edge]  (i1x) -- (i10);
  \draw[medge] (e78) -- (e785);
  \node[dL] at (1.15,1.45) {Circulatory};
  \node[dL] at (1.15,0.35) {Hypertensive dis.};
  \node[dL,text=blue!70!black] at (1.15,-0.85) {Ess.\ hypertension};
  \node[dR] at (4.85,1.45) {Endocrine \& metabolic};
  \node[dR] at (4.85,0.35) {Lipoprotein metab.};
  \node[dR,text=red!70!black] at (4.85,-0.85) {Hyperlipidemia};
  \node[font=\scriptsize\bfseries,text=blue!70!black,anchor=west] at (1.15,-1.35) {recorded (hit)};
  \node[font=\scriptsize\bfseries,text=red!70!black,anchor=west]  at (4.85,-1.35) {missed};
\node[lvl,anchor=south] at (3.0,1.25) {Chapter};
\node[lvl,anchor=south] at (3.0,0.15) {Section};
\node[lvl,anchor=south] at (3.0,-0.95) {Billable code};
  \end{tikzpicture}
  \caption{A real traversal from the top-50 test set (gold $=\{$I10, E785$\}$). From \textsc{root}
 the policy selects Chapter~9 (circulatory), enters \emph{Hypertensive diseases}, and records the
  billable leaf \textbf{I10} (essential hypertension), which is correct (blue). It never expands the
  endocrine/metabolic branch (red, dashed), so the gold code \textbf{E785} (hyperlipidemia) is missed:
  the comorbidity is in the coded record but never documented in the note body, illustrating that
  recall is bounded by what the narrative states.}
  \label{fig:traversal}
  \end{figure}

The environment enforces breadth-first traversal: each \texttt{expand\_level()} call processes every pending node at the same hierarchy depth in a single turn, so a complete traversal spans 5-7 depth levels (approximately 12--14 model turns) regardless of how many codes the note carries (Figure~\ref{fig:traversal}). 

The model interacts with the environment through Qwen3's function-calling interface. The rollout driver parses each \texttt{<tool\_call>}, executes the requested tool, and returns the result as a tool-response turn. During training, model-generated tokens are trainable and environment-injected tokens are masked; during evaluation, active rollouts are batched by wave for tractability at 1{,}000+ notes.

\subsection{Shared Traversal Policy (SFT-1)}
\label{sec:shared}
A single traversal policy makes all selection decisions across hierarchy depths within one continuous context window, conditioning each level on the accumulated traversal history. Prior LLM traversal of ICD hierarchies~\cite{boyle} relies on prompting alone; we train the policy directly on oracle demonstrations. We fine-tune on the 2{,}000-note SFT split using the gold trajectories. Each trajectory is replayed through the rollout driver to produce the trainable/masked split described in Section~\ref{sec:env}. All fine-tuning uses QLoRA~\cite{qlora} with rank $r{=}16$ on a 4-bit quantized base model; continuation arms warm-start from the preceding adapter.

\begin{figure}[t]
\centering
\begin{tikzpicture}[
  font=\tiny,
  box/.style={draw, rounded corners=2pt, align=center, inner sep=2pt, minimum height=5mm},
  adapter/.style={draw, fill=blue!8, rounded corners=2pt, align=center, inner sep=2pt,
                  minimum height=4.5mm, minimum width=11mm},
  envbox/.style={draw, fill=green!10, rounded corners=2pt, align=center, inner sep=2pt,
                 minimum height=5mm},
  >={Stealth[length=3pt]}, node distance=2.5mm]

\node[font=\tiny\bfseries] (title1) {Shared Traversal Policy};
\node[box, fill=gray!10, minimum width=28mm, below=2mm of title1] (note1)
     {Discharge note};
\node[box, fill=orange!12, minimum width=28mm, below=of note1] (bb1)
     {Qwen3-4B-Instruct-2507\\4-bit quantized};
\node[adapter, below=4mm of bb1] (a1) {QLoRA $r{=}16$\\(single adapter)};
\node[envbox, minimum width=28mm, below=4mm of a1] (env1)
     {Traversal env.\\\texttt{expand / select}};
\node[box, fill=gray!10, minimum width=28mm, below=of env1] (out1)
     {ICD-10-CM codes};

\draw[->] (note1) -- (bb1);
\draw[->] (bb1) -- (a1);
\draw[->] (a1) -- (env1);
\draw[->] (env1) -- (out1);
\draw[->, dashed] (env1.west) -- ++(-3mm,0) |- node[left, pos=0.25]
     {\tiny multi-turn} (a1.west);

\node[font=\tiny\bfseries, right=20mm of title1] (title2) {Cascade Baseline};
\node[box, fill=gray!10, minimum width=30mm, below=2mm of title2] (note2)
     {Discharge note};
\node[box, fill=orange!12, minimum width=30mm, below=of note2] (bb2)
     {Qwen3-4B-Instruct-2507\\4-bit quantized};
\node[adapter, below=4mm of bb2] (s2) {$A_2$\\subsect.};
\node[adapter, left=1mm of s2] (s1) {$A_1$\\chapter};
\node[adapter, right=1mm of s2] (s3) {$A_3$\\code};
\node[box, fill=gray!10, minimum width=30mm, below=5mm of s2] (out2)
     {ICD-10-CM codes};

\draw[->] (note2) -- (bb2);
\draw[->] (bb2.south) -- ++(0,-1.5mm) -| (s1.north);
\draw[->] (bb2.south) -- ++(0,-1.5mm) -| (s2.north);
\draw[->] (bb2.south) -- ++(0,-1.5mm) -| (s3.north);
\draw[->] (s1.south) -- ++(0,-1.5mm) -| (s2.west);
\draw[->] (s2.south) -- ++(0,-1.5mm) -| (s3.west);
\draw[->] (s3.south) -- ++(0,-1.5mm) -- ++(0,-1mm) -| (out2.north);

\draw[gray!40, dashed] ($(title1.east)!0.5!(title2.west)+(0,2mm)$) --
     ($(out1.east)!0.5!(out2.west)+(0,-2mm)$);
\end{tikzpicture}
\caption{Architecture comparison. \textbf{Left:} shared traversal policy (single adapter, multi-turn). \textbf{Right:} cascade baseline (three specialist adapters, single-turn each). Both share the same 4-bit quantized backbone.}
\label{fig:architecture}
\end{figure}
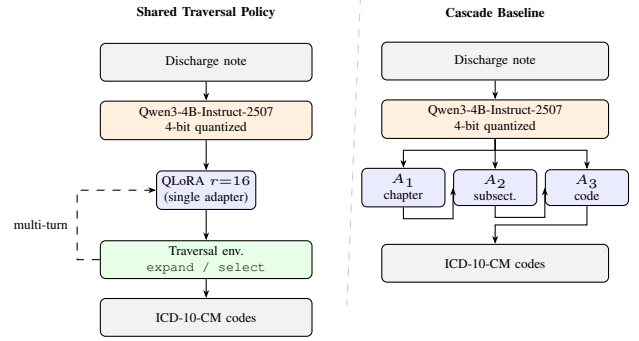

\subsection{Cascade Baseline (SFT-3)}
\label{sec:cascade}
The cascade decomposes the same traversal into three independent single-turn classification problems, one per hierarchy depth: a chapter specialist, a subsection specialist, and a code specialist. Each specialist receives the discharge note and a flat list of candidates at its assigned level, and responds with a single line (\texttt{SELECTED: id1, id2, ...}) choosing every candidate the note supports. There is no tool-call protocol, no multi-turn interaction, and no shared context across levels. At inference time the three are executed sequentially, so the chapter specialist's selections determine which candidates the subsection specialist sees, and in turn which billable codes are presented to the code specialist. All three share a common system prompt (Supplementary Materials, Appendix~A).

The candidate lists at the first two levels (L1: chapters, L2: subsections) are modest at both label-space scales, but the third level (L3: billable codes) presents all billable-leaf descendants of selected subsections in one batched prompt. This makes the cascade efficient, requiring three model calls per note, but creates a scaling limitation. Under the \emph{all} label space, the L3 prompt often exceeds the 32{,}000-token cap used for a single 46GB GPU. During training, over-cap L3 examples are dropped rather than truncated; during evaluation, the candidate tail is truncated, imposing a hard recall ceiling on 28\% of SFT-3 test notes and 32\% of SFT-3$^{+}$ test notes. The shared traversal policy has 0\% truncation at the same scale.

We train the three specialists on the same 2{,}000-note SFT split used for SFT-1, with each note re-expressed as three single-turn examples (one per level), yielding 6{,}000 training examples total.

\subsection{Continuation and RL Arms}
\label{sec:arms}
The shared traversal policy and cascade baseline each use 2{,}000 notes for initial training. To isolate the effects of data, architecture, and training algorithm, we define three continuation arms on the same 3{,}000-note continuation split.

\textbf{SFT-1$^{+}$} warm-starts from SFT-1 and continues supervised fine-tuning on 3{,}000 additional gold trajectories constructed as in Sections~\ref{sec:gold} and~\ref{sec:env}.

\textbf{SFT-3$^{+}$} warm-starts each SFT-3 specialist and continues supervised fine-tuning on the same continuation split, re-expressed as per-level single-turn examples.

\textbf{GRPO-1} warm-starts from SFT-1 and applies Group Relative Policy Optimization on the same 3{,}000 notes. The model generates live rollouts against the environment, and each rollout is scored with the terminal reward from Section~\ref{sec:mdp}. We run GRPO-1 on the \emph{frequent} label space only; the \emph{all} variant was omitted because training time scales to approximately 2--3 GPU-days and the \emph{frequent} result already shows no benefit over supervised continuation.

\subsection{Evaluation Protocol}
\label{sec:eval}
All arms are evaluated on the same held-out test set (1{,}000 notes) and validation set (500 notes), shared across both label spaces and all training arms (Section~\ref{sec:splits}). Evaluation rollouts are executed through the batched rollout driver (Section~\ref{sec:env}) under greedy decoding (temperature 0) for determinism and paired comparability.

\textbf{Zero-shot floors.} To anchor the fine-tuning results, we evaluate two zero-shot baselines using the base model base model (Qwen3-4B-Instruct-2507~\cite{qwen3}) with no task-specific training. The traversal-format base (Base in Table~\ref{tab:main}) receives the same system prompt, tool schema, and discharge note as SFT-1, scoring near-zero \microF{} on both label spaces (0.001), reflecting protocol-following failure rather than lack of clinical knowledge. The cascade base (Base-3) evaluates the same untrained model under the single-turn \texttt{SELECTED:} format of SFT-3, clearing the floor meaningfully (0.229 on \emph{frequent}, 0.058 on \emph{all}). The gap between the two floors quantifies the protocol-following overhead that the gold trajectories (Section~\ref{sec:gold}) are designed to close.

\textbf{Metrics.} We report micro-averaged precision, recall, and F1 (pooled across all notes and codes, weighting each prediction equally) and macro-averaged precision, recall, and F1 (averaged across the codes present in the gold set, weighting rare and common codes equally). Micro-F1 is the headline metric; macro-F1 surfaces rare-code performance, which dominates the \emph{all} label space. We additionally report path-F1, a hierarchical variant that awards partial credit for predictions landing in the correct region of the ICD-10-CM hierarchy rather than requiring an exact leaf match. For a code $c$, let $A(c) = \{c\} \cup \mathrm{anc}(c)$ denote its augmentation with all ancestors along the path toward the root (the root itself excluded), and extend this to a set of codes by $A(C) = \bigcup_{c \in C} A(c)$. Writing $\hat{C}_i$ and $C_i$ for the predicted and gold codes of note $i$, path-F1 replaces exact-match set membership with membership over the augmented sets:
{\small
\begin{equation}
\label{eq:path}
\text{path-P} = \frac{\sum_i |A(\hat{C}_i) \cap A(C_i)|}{\sum_i |A(\hat{C}_i)|}, \;\;
\text{path-R} = \frac{\sum_i |A(\hat{C}_i) \cap A(C_i)|}{\sum_i |A(C_i)|},
\end{equation}}
with path-F1 their harmonic mean. Equation~\ref{eq:path} gives the micro variant (sums pooled over notes); the macro variant averages the per-code F1 over the augmented gold codes. Because a mispredicted leaf still shares ancestors with the true code, this rewards a prediction that lands on the correct branch over one that is entirely unrelated.

\textbf{Statistical testing.} We use a paired bootstrap to assess differences between arms. Because micro-F1 and macro-F1 are pooled statistics that do not decompose into per-note scores, a paired t-test is not applicable. For each of 5{,}000 resamples we draw 1{,}000 test notes with replacement and recompute both arms' pooled metrics on the identical resampled set. We report the mean difference, 95\% confidence interval, and two-sided p-value (values below $1/5{,}000$ reported as $p < 0.001$). We report uncorrected p-values for approximately fifteen planned comparisons.\footnote{The large effects (supervised data) survive Bonferroni correction. The small architecture deltas that reach nominal significance ($p = 0.002$--$0.026$) would not all survive correction at $\alpha = 0.05/15 \approx 0.003$, and we interpret them accordingly.}

We additionally report per-level selection accuracy (chapter, subsection, billable code) to diagnose where recall is lost, and for the cascade under the \emph{all} label space, the fraction of test notes affected by L3 candidate-list truncation (Section~\ref{sec:cascade}).

\section{Results}
\label{sec:results}
\subsection{Overall Performance}
Table~\ref{tab:main} reports micro- and macro-averaged precision, recall, and F1 for all arms on the held-out test set ($n{=}1{,}000$) at both label-space scales. The results are summarized below.

\begin{table*}[t]
\caption{Micro- and macro-averaged performance on the held-out test set ($n{=}1{,}000$) by training arm and label space. CAML, LAAT, and PLM-ICD are flat multi-label baselines retrained on the same splits. Path-F1 gives hierarchical partial credit (Section~\ref{sec:eval}).}
\label{tab:main}
\centering
\begin{tabular}{llcccccccc}
\toprule
& & \multicolumn{4}{c}{Micro-averaged} & \multicolumn{4}{c}{Macro-averaged} \\
\cmidrule(lr){3-6} \cmidrule(lr){7-10}
Variant & Arm & P & R & F1 & path-F1 & P & R & F1 & path-F1 \\
\midrule
Frequent
 & Base (0-shot)   & .0010 & .0017 & .0013 & .0062 & .0650 & .0014 & .0027 & .0041 \\
 & Base-3 (0-shot) & .4013 & .1597 & .2285 & .2612 & .3499 & .1478 & .1911 & .2009 \\
\cmidrule(lr){2-10}
 & CAML             & .6576 & .6778 & .6675 & ---   & .5738 & .6040 & .5819 & ---   \\
 & LAAT             & .6675 & .7250 & .6951 & ---   & .5944 & .6552 & .6157 & ---   \\
 & PLM-ICD          & .6864 & .6962 & .6912 & ---   & .5928 & .6488 & .6125 & ---   \\
\cmidrule(lr){2-10}
 & SFT-1           & .6734 & .7020 & .6874 & .7146 & .6210 & .6425 & .6229 & .6347 \\
 & SFT-1$^{+}$     & .7150 & .7031 & .7090 & .7379 & .6719 & .6529 & .6514 & .6639 \\
 & GRPO-1          & .6558 & .7230 & .6878 & .7138 & .5958 & .6568 & .6174 & .6283 \\
 & SFT-3           & .6751 & .7037 & .6891 & .7190 & .6267 & .6471 & .6299 & .6466 \\
 & SFT-3$^{+}$     & .7048 & .7008 & .7028 & .7358 & .6625 & .6422 & .6394 & .6620 \\
\midrule
All
 & Base (0-shot)   & .0018 & .0009 & .0012 & .0090 & .0029 & .0009 & .0010 & .0028 \\
 & Base-3 (0-shot) & .0701 & .0488 & .0575 & .1337 & .0806 & .0694 & .0596 & .0795 \\
\cmidrule(lr){2-10}
 & CAML             & .1901 & .1976 & .1938 & ---   & .0011 & .0057 & .0018 & ---   \\
 & LAAT             & .5255 & .4466 & .4829 & ---   & .1026 & .1157 & .0988 & ---   \\
 & PLM-ICD          & .1921 & .1996 & .1958 & ---   & .0011 & .0057 & .0018 & ---   \\
\cmidrule(lr){2-10}
 & SFT-1           & .5217 & .4402 & .4775 & .5784 & .2473 & .2198 & .2172 & .2576 \\
 & SFT-1$^{+}$     & .5653 & .4928 & .5266 & .6225 & .2828 & .2610 & .2562 & .2936 \\
 & SFT-3           & .5024 & .4505 & .4750 & .5800 & .2475 & .2385 & .2269 & .2652 \\
 & SFT-3$^{+}$     & .5388 & .4889 & .5126 & .6106 & .2823 & .2724 & .2602 & .2992 \\
\bottomrule
\end{tabular}
\end{table*}

\textbf{The traversal policy outperforms flat baselines, with the margin growing at scale.} Three flat multi-label baselines anchor these results externally. CAML and PLM-ICD both reach competitive \microF{} on \emph{frequent} (0.668 and 0.691) but collapse to near-identical scores on \emph{all} (0.194 and 0.196), with macro-F1 falling to 0.002 for both---indicating that neither convolutional attention nor pretrained transformer encoders can discriminate among 15{,}761 candidates without structural guidance. Only LAAT survives at scale (0.483 on \emph{all}), and SFT-1 trained on just 2{,}000 trajectories already matches it (0.478 vs.\ 0.483). SFT-1$^{+}$ significantly outperforms all three baselines at both scales (Table~\ref{tab:boot}), with the margin widening sharply on \emph{all}: $+0.044$ \microF{} over LAAT and $+0.157$ \macroF{}, confirming that the traversal formulation's advantage is concentrated where flat scoring is weakest. Notably, PLM-ICD's pretrained encoder offers no advantage over CAML's CNN at the full scale, suggesting that the bottleneck is the flat output layer rather than the representation - with only 2{,}000 training notes the vast majority of the 15{,}761 output neurons see fewer than five positive examples.

\textbf{More supervised data helps, and the effect grows with scale.} The $+3{,}000$-note continuation (SFT-1$^{+}$ over SFT-1) is the largest gain in the table: $+0.022$ \microF{} on \emph{frequent} ($0.687 \rightarrow 0.709$) and $+0.049$ on \emph{all} ($0.478 \rightarrow 0.527$). The lift roughly doubles from the 50-code to the full label space. SFT-1$^{+}$ is the best arm in both settings, and the same pattern holds for the cascade (SFT-3$^{+}$ over SFT-3 by comparable margins). Macro-F1 gains track micro, so the improvement is not confined to high-frequency codes.

\textbf{GRPO does not improve over supervised continuation.} GRPO-1 (0.688 \microF{}) ties its SFT-1 warm-start (0.687) and falls below SFT-1$^{+}$ (0.709) on both micro and macro. Training directly on the coding metric via RL on 3{,}000 notes offers no benefit over simply continuing supervised fine-tuning on the same data from the same initialization.

\textbf{Neither architecture wins consistently.} The cascade (SFT-3, SFT-3$^{+}$) lands within roughly one \microF{} point of the corresponding shared-policy arm (SFT-1, SFT-1$^{+}$) at both scales. On \emph{all}, SFT-3$^{+}$ scores 0.513 vs.\ SFT-1$^{+}$'s 0.527 on micro but 0.260 vs.\ 0.256 on macro; the deltas flip sign across metrics and variants. Statistical tests follow in Section~\ref{sec:stats}. Where the cascade does lose is structurally: 28--32\% of test notes suffer L3 candidate-list truncation under the \emph{all} label space (Section~\ref{sec:cascade}), a constraint the shared policy never encounters.

\subsection{Statistical Comparisons}
\label{sec:stats}
Table~\ref{tab:boot} reports paired-bootstrap estimates (5{,}000 resamples) of the difference in pooled micro- and macro-F1 between matched arms, with 95\% confidence intervals and two-sided p-values. All external comparisons are significant ($p < 0.001$). The margins are modest on \emph{frequent} ($+0.014$ to $+0.042$ \microF{}) but large on \emph{all} ($+0.044$ to $+0.333$ \microF{}), confirming that the traversal formulation's advantage grows with label-space scale. On \emph{frequent}, SFT-1 without continuation data already matches LAAT ($p = 0.08$, n.s.).

Among the internal comparisons, the data effect is the only robust finding: SFT-1$^{+}$ over SFT-1 is significant on all four tests ($p < 0.001$), with the lift doubling from \emph{frequent} to \emph{all}. GRPO-1 ties SFT-1 (micro $p = 0.892$) and falls significantly below SFT-1$^{+}$ ($p < 0.001$). Architecture deltas are small ($\leq 0.014$), inconsistent in sign, and would not survive Bonferroni correction.

\begin{table*}[t]
\caption{Paired bootstrap comparisons (5{,}000 resamples): $\Delta$ micro- and macro-F1 with $p$-values. External baselines (CAML, LAAT, PLM-ICD) above the rule; internal architecture, algorithm, and data comparisons below.}
\label{tab:boot}
\centering
\setlength{\tabcolsep}{5pt}
\begin{tabular}{llcccccc}
\toprule
Variant & Comparison & Metric & New & Base & $\Delta$ & $p$ & Verdict \\
\midrule
Frequent
& SFT-1$^{+}$ vs CAML    & Micro & 0.7090 & 0.6675 & $+0.042$ & $<0.001$ & $\uparrow$ higher \\
 &                        & Macro & 0.6514 & 0.5819 & $+0.070$ & $<0.001$ & $\uparrow$ higher \\
 & SFT-1$^{+}$ vs LAAT    & Micro & 0.7090 & 0.6951 & $+0.014$ & $<0.001$ & $\uparrow$ higher \\
 &                        & Macro & 0.6514 & 0.6157 & $+0.036$ & $<0.001$ & $\uparrow$ higher \\
 & SFT-1$^{+}$ vs PLM-ICD & Micro & 0.7090 & 0.6912 & $+0.018$ & $<0.001$ & $\uparrow$ higher \\
 &                        & Macro & 0.6514 & 0.6125 & $+0.039$ & $<0.001$ & $\uparrow$ higher \\
 & SFT-1 vs LAAT          & Micro & 0.6874 & 0.6951 & $-0.008$ & $0.084$ & n.s. \\
 &                        & Macro & 0.6229 & 0.6157 & $+0.007$ & $0.180$ & n.s. \\
\cmidrule(lr){2-8}
 & SFT-1$^{+}$ vs SFT-1  & Micro & 0.7090 & 0.6874 & $+0.022$ & $<0.001$ & $\uparrow$ higher \\
 &                        & Macro & 0.6514 & 0.6229 & $+0.029$ & $<0.001$ & $\uparrow$ higher \\
 & GRPO-1 vs SFT-1        & Micro & 0.6878 & 0.6874 & $+0.000$ & $0.892$ & n.s. \\
 &                        & Macro & 0.6174 & 0.6229 & $-0.006$ & $0.104$ & n.s. \\
 & GRPO-1 vs SFT-1$^{+}$  & Micro & 0.6878 & 0.7090 & $-0.021$ & $<0.001$ & $\downarrow$ lower \\
 &                        & Macro & 0.6174 & 0.6514 & $-0.034$ & $<0.001$ & $\downarrow$ lower \\
 & SFT-3 vs SFT-1         & Micro & 0.6891 & 0.6874 & $+0.002$ & $0.638$ & n.s. \\
 &                        & Macro & 0.6299 & 0.6229 & $+0.007$ & $0.142$ & n.s. \\
 & SFT-3$^{+}$ vs SFT-1$^{+}$ & Micro & 0.7028 & 0.7090 & $-0.006$ & $0.086$ & n.s. \\
 &                        & Macro & 0.6394 & 0.6514 & $-0.012$ & $0.017$ & $\downarrow$ lower \\
\midrule
All
 & SFT-1$^{+}$ vs CAML    & Micro & 0.5266 & 0.1938 & $+0.333$ & $<0.001$ & $\uparrow$ higher \\
 &                        & Macro & 0.2562 & 0.0018 & $+0.254$ & $<0.001$ & $\uparrow$ higher \\
 & SFT-1$^{+}$ vs LAAT    & Micro & 0.5266 & 0.4829 & $+0.044$ & $<0.001$ & $\uparrow$ higher \\
 &                        & Macro & 0.2562 & 0.0988 & $+0.157$ & $<0.001$ & $\uparrow$ higher \\
 & SFT-1$^{+}$ vs PLM-ICD & Micro & 0.5266 & 0.1958 & $+0.331$ & $<0.001$ & $\uparrow$ higher \\
 &                        & Macro & 0.2562 & 0.0018 & $+0.254$ & $<0.001$ & $\uparrow$ higher \\
\cmidrule(lr){2-8}
  & SFT-1$^{+}$ vs SFT-1  & Micro & 0.5266 & 0.4775 & $+0.049$ & $<0.001$ & $\uparrow$ higher \\
 &                        & Macro & 0.2562 & 0.2172 & $+0.039$ & $<0.001$ & $\uparrow$ higher \\
 & SFT-3 vs SFT-1         & Micro & 0.4750 & 0.4775 & $-0.003$ & $0.598$ & n.s. \\
 &                        & Macro & 0.2269 & 0.2172 & $+0.010$ & $0.026$ & $\uparrow$ higher \\
 & SFT-3$^{+}$ vs SFT-1$^{+}$ & Micro & 0.5126 & 0.5266 & $-0.014$ & $0.002$ & $\downarrow$ lower \\
 &                        & Macro & 0.2602 & 0.2562 & $+0.004$ & $0.545$ & n.s. \\
 
\bottomrule
\end{tabular}
\end{table*}

\subsection{Hierarchical Recall by Depth}
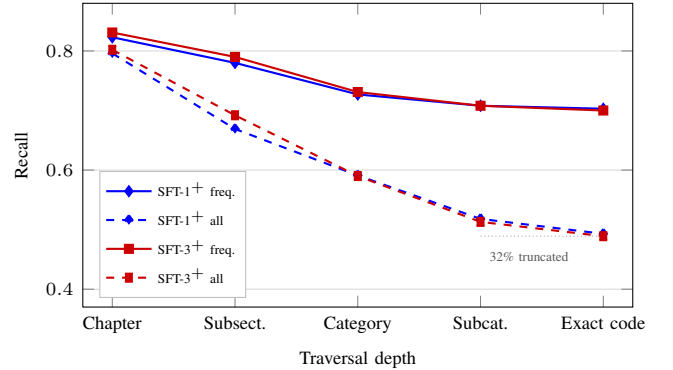
\begin{figure}[t]
\centering
\begin{tikzpicture}
\begin{axis}[
  width=\columnwidth, height=5.6cm,
  symbolic x coords={Chapter, Subsect., Category, Subcat., Exact code},
  xtick=data,
  x tick label style={font=\scriptsize},
  xlabel={Traversal depth}, xlabel style={font=\scriptsize},
  ylabel={Recall}, ylabel style={font=\scriptsize},
  ymin=0.37, ymax=0.88,
  ytick={0.4,0.6,0.8},
  y tick label style={font=\scriptsize},
  ymajorgrids, grid style={gray!30},
  legend style={at={(0.03,0.03)}, anchor=south west, font=\tiny,
    cells={anchor=west}, draw=gray!50, fill=white, inner sep=2pt},
  legend cell align=left,
  enlarge x limits=0.06,
]
\addplot[blue, thick, mark=diamond*, mark size=1.8pt]
  coordinates {(Chapter,0.823) (Subsect.,0.780) (Category,0.727) (Subcat.,0.708) (Exact code,0.703)};
\addlegendentry{SFT-1$^{+}$ freq.}
\addplot[blue, thick, dashed, mark=diamond*, mark size=1.8pt]
  coordinates {(Chapter,0.796) (Subsect.,0.669) (Category,0.591) (Subcat.,0.518) (Exact code,0.493)};
\addlegendentry{SFT-1$^{+}$ all}
\addplot[red!80!black, thick, mark=square*, mark size=1.5pt]
  coordinates {(Chapter,0.831) (Subsect.,0.790) (Category,0.731) (Subcat.,0.708) (Exact code,0.700)};
\addlegendentry{SFT-3$^{+}$ freq.}
\addplot[red!80!black, thick, dashed, mark=square*, mark size=1.5pt]
  coordinates {(Chapter,0.802) (Subsect.,0.692) (Category,0.590) (Subcat.,0.513) (Exact code,0.489)};
\addlegendentry{SFT-3$^{+}$ all}
\node[font=\tiny, anchor=west, text=gray!70!black]
  at (axis cs:Subcat.,0.455) {32\% truncated};
\draw[gray!70, densely dotted]
  (axis cs:Subcat.,0.489) -- (axis cs:Exact code,0.489);
\end{axis}
\end{tikzpicture}
\caption{Micro-pooled recall by traversal depth for SFT-1$^{+}$ (blue) and SFT-3$^{+}$ (red) under \emph{frequent} (solid) and \emph{all} (dashed). At the \emph{all} scale, 32\% of cascade test notes suffer L3 truncation. Depths d5-d6 omitted (ragged paths).}
\label{fig:depth}
\end{figure}

To diagnose where in the hierarchy codes are lost, we compute micro-pooled recall at each depth of the ICD-10-CM tree: at depth $d$, each predicted and gold leaf code is collapsed to its depth-$d$ ancestor, and recall is computed over the resulting sets.\footnote{Codes whose paths are shorter than $d$ do not contribute at that depth, so deeper columns reflect only codes that actually reach that level of the hierarchy. The denominators therefore shrink at deeper levels.} Figure~\ref{fig:depth} plots this for the best shared-policy and cascade arms at both label-space scales.

Recall degrades monotonically with depth, from 0.80-0.83 at the chapter level to 0.70 (frequent) and 0.49 (all) at the billable leaf. The \emph{frequent} $\rightarrow$ \emph{all} gap widens at each depth, confirming that rare-code discrimination at the bottom of the tree drives the \macroF{} collapse. The cascade tracks the shared policy at shallow depths but diverges where L3 truncation (32\% of test notes) imposes a hard recall ceiling the shared policy never encounters.

\section{Discussion}

\subsection{The Rare-Code Bottleneck}
The \macroF{} collapse from \emph{frequent} to \emph{all}
($0.651 \rightarrow 0.256$ for the best arm) is the starkest result in the study. Micro-F1 also drops ($0.709 \rightarrow 0.527$), but the far larger macro gap shows the degradation is concentrated in the long tail of rare codes rather than spread uniformly. The flat baselines degrade far more sharply, confirming that the traversal formulation's hierarchical decomposition partially insulates against the rare-code problem without solving it. The per-depth analysis (Figure~\ref{fig:depth}) corroborates this: chapter-level recall is comparable across scales, but the gap widens at each depth, largest at the billable leaf where rare codes dominate.

\subsection{Supervision Is the Bottleneck}
Of the three variables we isolate, only additional supervised data reliably improves performance (Table~\ref{tab:boot}). Neither RL nor architectural specialization produces a comparable or even consistent gain.
This is worth stating plainly because the natural design impulse points the other way: once a hierarchy is available, it is tempting to assign specialist models to each level and to replace hand-built trajectories with learned rewards. Our factorial study tests both assumptions and finds neither justified at this scale. Our results suggest that, at least for extraction-bound graph-constrained coding at this model scale, the complexity does not pay off. GRPO ties its supervised warm-start and loses to supervised continuation on the same data; the cascade matches the shared policy but never beats it. The bottleneck is not how the model is trained or how the traversal is decomposed, but how many gold demonstrations it sees.

This conclusion is scoped to the regime we test (a single 4B model, one reward design, 2{,}000--5{,}000 notes), and RL or specialization may yet help at larger scales or with richer reward shaping. What our study contributes is a controlled baseline: any added machinery should be shown to beat simply adding more supervised data at matched compute.

\subsection{Consolidation Scales Where the Cascade Cannot}
On headline accuracy the shared traversal policy and the three-specialist cascade are statistically indistinguishable at both label-space scales. The consolidation argument therefore rests not on accuracy but on deployment simplicity (one adapter, no inter-stage orchestration) and a structural scaling advantage.

The cascade hits a scaling wall at the full code space: its batched L3 prompt forces every billable-leaf descendant of the selected subsections into one context window, overflowing it for 28-32\% of test notes and making those codes unselectable regardless of model quality. A larger memory budget would raise this ceiling, but the pressure is structural - the single-shot design's prompt cost grows with label-space size, whereas the shared policy's per-turn fan-out distributes the same load across turns (0\% truncation).

\subsection{Limitations}
All results are from a single 4B-parameter model (Qwen3-4B-Instruct-2507); RL and architectural specialization may behave differently at larger scales or with other model families. The GRPO result is based on a single reward design, one hyperparameter configuration, and one warm-start on the \emph{frequent} label space only; the full-scale RL cell is argued but not directly measured. All data comes from MIMIC-IV, a single US academic medical center, and the gold labels are billed administrative codes known to be incomplete, so measured precision is a lower bound when the model assigns a clinically valid code that was not billed. The 32{,}000-token context cap driving the cascade's 28-32\% truncation rate is a memory budget constraint (single 46GB GPU), not an architectural limit of the model. Finally, our training budget of 5{,}000 notes is small relative to the 122{,}197 available; the data-scaling claim rests on just two training budgets (2{,}000 and 5{,}000 notes) and we do not determine where the lift saturates.

\section{Conclusion}
We formulated ICD-10-CM code prediction as a single graph-constrained traversal policy that significantly outperforms flat
multi-label baselines at both label-space scales (by $+0.044$
\microF{} and $+0.157$ \macroF{} over the strongest (LAAT) on
the full 15{,}761-code space) and conducted a controlled study
isolating architecture, training algorithm, and data. Across both scales, only additional supervised data reliably helped, RL offered no benefit over supervised continuation, and one shared policy matched the three-specialist cascade while avoiding its context-window truncation. Practitioners should therefore deploy one model, not three, and invest in annotation rather than architectural or algorithmic complexity.

Several directions remain open. The data-scaling curve has only two budgets (2{,}000 and 5{,}000 notes); larger annotation budgets and targeted sampling of rare-code-rich notes could determine where the supervised lift saturates. Sampling-based decoding (self-consistency, best-of-$n$) and per-code threshold tuning may improve rare-code recall, which the \macroF{} collapse identifies as the primary bottleneck. RL deserves re-examination with richer reward shaping and at larger model scales before a general negative conclusion is warranted. Finally, validation on a second institution and coding system would test whether these findings generalize beyond a single US academic center.

\section*{Acknowledgment}
 This material is based upon work supported by the National Science Foundation under Grant No. 2431840.

\newpage
\input{supplementarymaterials}

\end{document}

%% file: supplementarymaterials.tex
\clearpage
\appendices

\section*{Supplementary Materials}
\addcontentsline{toc}{section}{Supplementary Materials}

\section{System Prompts}
\label{app:prompts}

\subsection{Shared Agentic Policy (SFT-1 / SFT-1\textsuperscript{+} / GRPO-1)}
\label{app:prompt_sft1}

The following system prompt is used for the shared agentic policy across all training and evaluation conditions. The model additionally receives the tool schema for \texttt{expand\_level()} and \texttt{select(children)} as function definitions in Qwen3's native tool-calling format.

\begin{lstlisting}
Assign ICD-10-CM codes by traversing the ontology
level-by-level using tools.

Rules:
- ONLY select IDs explicitly listed in the most
  recent expand_level() output.
- Do NOT guess codes or use external knowledge of
  the ICD hierarchy.
- Only the most specific (leaf) codes are billable
  -- drill as deep as the note supports.

Process:
1. Call expand_level() -- shows all pending nodes
   and their children, grouped by parent. Leaf codes
   (no sub-codes exist) are auto-recorded immediately.
2. Call select(children=[...]) with the IDs from ANY
   group that the note supports. Pass an empty list
   if none of the current options apply.
3. Repeat until expand_level() returns DONE.
4. Output exactly: ANSWER: code1, code2, ...

Think in 1 SHORT sentence, then immediately call
the tool.
\end{lstlisting}

\subsubsection{Tool Schema}
The model receives the following tool definitions in Qwen3's native
tool-calling format.

\begin{lstlisting}
{
  "type": "function",
  "function": {
    "name": "expand_level",
    "description": "Show all pending nodes and their
      children grouped by parent. Leaf codes (no
      sub-codes) are auto-recorded as final answers.
      Returns DONE when the agenda is empty.",
    "parameters": {
      "type": "object",
      "properties": {},
      "required": []
    }
  }
}

{
  "type": "function",
  "function": {
    "name": "select",
    "description": "Queue child node IDs for further
      expansion. Only leaf codes can become final
      answers (auto-recorded by expand_level).
      Non-leaf codes must be drilled into -- they
      cannot be recorded directly. Pass children=[]
      to skip all current options if none apply to
      the note.",
    "parameters": {
      "type": "object",
      "properties": {
        "children": {
          "type": "array",
          "items": {"type": "string"},
          "description": "Child node IDs to drill
            into. Pass an empty list to skip all
            options at this level."
        }
      },
      "required": ["children"]
    }
  }
}
\end{lstlisting}

\subsubsection{Example Trajectory}

The user message contains only the clinical note. The following illustrates the multi-turn tool-call interaction for a note with gold codes E785, I10, and J449. A complete traversal typically spans 5-7 depth levels, each requiring an expand-and-select round trip (approximately 12-14 model turns total).

\begin{lstlisting}
[assistant]
<tool_call>{"name": "expand_level",
  "arguments": {}}</tool_call>

[tool response]
Under ROOT: Ch3 (Diseases of the blood ...),
Ch4 (Endocrine ... (E00-E89)),
Ch9 (Circulatory ... (I00-I99)),
Ch10 (Respiratory ... (J00-J99)), ... Ch21 (...)
Call select(children=[...]) with IDs from ANY
group above.

[assistant]
<tool_call>{"name": "select",
  "arguments": {"children":
    ["Ch4", "Ch9", "Ch10"]}}</tool_call>

[tool response]
Selected ['Ch4', 'Ch9', 'Ch10'].
Call expand_level() to drill deeper.

  ... expand_level / select repeats at each depth,
  drilling each branch to its leaf ...

  (Ch4 -> E70-E88 -> E785 auto-recorded;
   Ch9 -> I10-I1A -> I10 auto-recorded;
   Ch10 -> J40-J4A -> J44 -> J449 auto-recorded)

[assistant]
ANSWER: I10, E785, J449
\end{lstlisting}

\subsection{Cascade Baseline (SFT-3 / SFT-3\textsuperscript{+})}
\label{app:prompt_sft3}

All three cascade specialists share the following system prompt. Each specialist receives a single-turn user message containing the clinical note and a candidate list specific to its ontology level. There is no tool-call protocol; the model responds with a single \texttt{SELECTED:} line.

\begin{lstlisting}
You are an expert ICD-10-CM coder. You assign codes
by navigating the ICD-10-CM ontology one level at a
time. Given a clinical note and a list of candidate
nodes at the current level (grouped by their parent),
select EVERY candidate the note supports.

Rules:
- ONLY choose IDs from the candidate list below.
  Do NOT invent codes or use outside knowledge
  of the hierarchy.
- Use the IDs verbatim.
- Reply with ONE line exactly:
  SELECTED: id1, id2, ...

If none of the candidates apply to the note, reply:
  SELECTED: NONE
\end{lstlisting}

\subsubsection{L1: Chapter Specialist}

User message format:

\begin{lstlisting}
Clinical note:
<medical_note>

Position: top of the ICD-10-CM hierarchy (chapters).
Candidates (choose all that apply):
- Ch3 (Diseases of the blood ... (D50-D89))
- Ch4 (Endocrine, nutritional and metabolic
  diseases (E00-E89))
- Ch9 (Diseases of the circulatory system
  (I00-I99))
  ...

Expected response:
SELECTED: Ch4, Ch9, Ch10
\end{lstlisting}

\subsubsection{L2: Subsection Specialist}

User message format:

\begin{lstlisting}
Clinical note:
<medical_note>

Selected chapters: Ch4 (Endocrine ... (E00-E89)),
Ch9 (Circulatory ... (I00-I99)),
Ch10 (Respiratory ... (J00-J99))

Candidates (choose all that apply):
Under Ch4 (Endocrine ...):
  - E00-E07 (Disorders of thyroid gland)
  - E08-E13 (Diabetes mellitus)
  - E70-E88 (Metabolic disorders)
Under Ch9 (Diseases of the circulatory system):
  - I10-I1A (Hypertensive diseases)
  - I20-I25 (Ischemic heart diseases)
  ...

Expected response:
SELECTED: E70-E88, I10-I1A, J40-J4A
\end{lstlisting}

\subsubsection{L3: Code Specialist}

User message format:

\begin{lstlisting}
Clinical note:
<medical_note>

Selected subsections: E70-E88 (Metabolic disorders),
I10-I1A (Hypertensive diseases),
J40-J4A (Chronic lower respiratory diseases)

Candidates (choose all that apply):
Under E70-E88 (Metabolic disorders):
  - E785 (Hyperlipidemia, unspecified)
  - E871 (Hypo-osmolality and hyponatremia)
  - E875 (Hyperkalemia)
Under I10-I1A (Hypertensive diseases):
  - I10 (Essential (primary) hypertension)
  - I110 (Hypertensive heart disease w/ HF)
  ...

Expected response:
SELECTED: E785, I10, J449
\end{lstlisting}

Note: under the \textit{all} label space, the L3 candidate list can reach tens of thousands of tokens (average $\approx$26k, maximum $\approx$74k). The sequence length is capped at 32,000 tokens to fit within the memory budget of a single 46GB GPU; the model itself supports up to 262k positions. See Section~IV-D of the main text for truncation handling.

\section{Rollout Driver Implementation}
\label{app:rollout}

\subsection{Overview}

The traversal environment requires a multi-turn loop: the model generates a tool call, the environment executes it and returns a result, and the model generates again conditioned on the full history. The training framework (TRL 0.22.2 with Unsloth) issues a single generation call per rollout and provides no native tool-calling loop. We implement the multi-turn cycle in two drivers that share identical per-rollout stepping logic.

\subsection{Single-Note Driver}

Used for tokenizing gold trajectories, GRPO training (per-prompt sequential rollouts), and single-note debugging. Each turn, the driver:

\begin{enumerate}
\item Calls the model (or replays a gold action during SFT data construction).
\item Parses the output for a \texttt{<tool\_call>} block using a regex pattern matching Qwen3's format: \texttt{<tool\_call>\{...\}</tool\_call>}.
\item Executes the parsed tool call against the environment (\texttt{expand\_level()} or \texttt{select(children=[...])}).
\item Injects the tool result as the next turn, wrapped in Qwen3's \texttt{<tool\_response>} format.
\end{enumerate}

Rather than re-rendering the full message list through the chat template at each turn (which would not round-trip correctly for Qwen3's \texttt{<think>} and \texttt{<tool\_call>} spans), the driver maintains a running concatenation of raw token IDs, appending each new segment directly. This produces the trainable/masked split by construction:

\begin{itemize}
\item \textbf{mask = 1 (trainable):} all tokens the model generated (tool-call content, reasoning, \texttt{ANSWER} line).
\item \textbf{mask = 0 (masked):} all tokens the driver injected (tool-result content, role headers such as \texttt{<|im\_start|>user} and \texttt{<|im\_start|>assistant}, \texttt{<tool\_response>} wrappers, and any \texttt{<think>} block opening tokens auto-inserted by the generation prompt).
\end{itemize}

\subsection{Batched Driver}

Used for evaluation only (not for GRPO training, which uses the sequential single-note driver). The per-rollout state machine is a line-for-line port of the single-note driver, verified byte-identical on a batch of one against a deterministic mock. The difference is operational: all active rollouts are pooled into a single \texttt{llm.generate()} call per wave. At each turn the driver:

\begin{enumerate}
\item Collects the running token context and remaining budget from every rollout that has not yet finished.
\item Issues one batched generation call with per-request \texttt{SamplingParams} (each rollout carries its own \texttt{max\_tokens} reflecting its remaining budget).
\item Steps each rollout individually: parsing its tool call, executing against its own environment instance, injecting its tool result.
\item Retires rollouts that emit \texttt{ANSWER} or exhaust their token budget.
\end{enumerate}

The total number of generation calls equals the maximum number of turns across all rollouts (capped at 16), not the sum of per-rollout turns. This is what makes multi-turn evaluation tractable at 1,000 test notes.

\subsection{Agentic Guardrails}

Two mechanisms prevent degenerate rollout behavior:

\textbf{Forced first expansion.} When enabled, the driver injects a masked \texttt{expand\_level()} call as the first assistant turn before the model generates anything. This seeds the loop with the chapter-level candidate list so the model cannot skip the traversal and emit a memorized \texttt{ANSWER} from the system prompt alone --- a failure mode observed in early GRPO runs where the model learned to terminate immediately to avoid negative reward.

\textbf{Early-answer rejection.} When enabled, if the model emits \texttt{ANSWER} before the environment has reported \texttt{DONE} (i.e., the agenda is not yet empty), the driver injects a masked user-turn nudge instructing the model to continue traversing:

\begin{lstlisting}
You have not finished traversing the ontology.
Do NOT output ANSWER yet. Continue by calling
select(children=[...]) or expand_level(). You may
only output ANSWER after expand_level() reports DONE.
\end{lstlisting}

The rollout remains active for the next wave. This prevents the model from terminating mid-traversal to collect a partial reward during GRPO, or from short-circuiting the traversal during evaluation.

\subsection{Qwen3 Framing Detection}

The injected text between turns (closing the assistant turn, wrapping the tool result, re-opening the next assistant turn) must match the chat template exactly. Qwen3-Instruct and Qwen3-Thinking variants differ in whether the generation prompt auto-opens a \texttt{<think>} block. At initialization, the driver probes the tokenizer with a minimal message list and inspects whether the rendered generation prompt ends with \texttt{<think>}. It then builds the inter-turn framing template accordingly:

\begin{lstlisting}
# Template for injecting tool results between turns:
<|im_start|>user
<tool_response>
{result}
</tool_response><|im_end|>
<|im_start|>assistant
<think>          # only if model auto-opens <think>
\end{lstlisting}

This auto-detection ensures the driver works with both Instruct and Thinking variants without manual configuration.

\section{Training Hyperparameters}
\label{app:hyperparams}

All arms use the same base model (Qwen3-4B-Instruct-2507) with 4-bit quantization and QLoRA adapters. The data-splitting seed is 42; the training seed is 3407. Table~\ref{tab:hyperparams_sft} reports the SFT configuration shared across SFT-1, SFT-1\textsuperscript{+}, SFT-3, and SFT-3\textsuperscript{+}. Table~\ref{tab:hyperparams_grpo} reports the GRPO-1 configuration.

\begin{table}[h]
\caption{QLoRA and SFT training configuration (shared across all SFT arms).}
\label{tab:hyperparams_sft}
\centering
\footnotesize
\begin{tabular}{@{}ll@{}}
\toprule
Parameter & Value \\
\midrule
Base model & Qwen3-4B-Instruct-2507 \\
Quantization & 4-bit (NF4) \\
LoRA rank ($r$) & 16 \\
LoRA alpha & 32 \\
LoRA dropout & 0 \\
Target modules & q/k/v/o/gate/up/down\_proj \\
Optimizer & AdamW (8-bit) \\
Learning rate & 2e-4 \\
LR schedule & Cosine \\
Warmup ratio & 0.05 \\
Epochs & 3 (best checkpoint by val loss) \\
Per-device batch size & 1 \\
Gradient accumulation & 8 (effective batch = 8) \\
Precision & bf16 \\
Max sequence length (\textit{frequent}) & 12,592 \\
Max sequence length (\textit{all}, agentic) & 20,000 \\
Max sequence length (\textit{all}, cascade L3) & 32,000 \\
Max completion length (\textit{frequent}) & 6,500 \\
Max completion length (\textit{all}, agentic) & 12,000 \\
Training seed & 3407 \\
\bottomrule
\end{tabular}
\end{table}

\begin{table}[h]
\caption{GRPO-1 training configuration. LoRA and base model settings are identical to Table~\ref{tab:hyperparams_sft}; only the training recipe differs.}
\label{tab:hyperparams_grpo}
\centering
\footnotesize
\begin{tabular}{@{}ll@{}}
\toprule
Parameter & Value \\
\midrule
Learning rate & 2e-5 \\
LR schedule & Linear decay \\
Warmup ratio & 0.03 \\
Epochs & 1 \\
Generations per prompt ($G$) & 4 \\
Sampling temperature & 1.0 \\
KL coefficient ($\beta$) & 0.0 (no KL penalty) \\
Per-device batch size & 1 \\
Gradient accumulation & 4 \\
Weight decay & 0.01 \\
Reward & code set-$F_1$ + path-$F_1$ (equal weight) \\
Rollout driver & Sequential (per-prompt) \\
Decoding (evaluation) & Greedy (temperature 0) \\
\bottomrule
\end{tabular}
\end{table}

The SFT and GRPO recipes are deliberately not matched on learning rate or epoch count: an SFT epoch (teacher-forced pass over gold trajectories) and a GRPO epoch (rollout, score, and policy update with $G$ generations per prompt) are different objectives with different optima. Fairness is ensured by the data-matched splits and the SFT-1\textsuperscript{+} compute control, not by identical hyperparameters across algorithms.

The \textit{all} variant uses larger sequence and completion budgets because gold traversals over the full 15,761-leaf ontology produce 3--4$\times$ longer token sequences. The \textit{frequent} budgets are non-binding for that variant. The cascade's L3 sequence cap of 32,000 tokens is a memory budget constraint for the single 46GB GPU used in this study; the base model (Qwen3-4B-Instruct-2507) natively supports up to 262,144 positions. The completion length budget does not apply to the cascade, whose specialists produce only a short single-turn \texttt{SELECTED:} response.

\section{Flat Baseline Configurations}
\label{app:baselines}

CAML, LAAT, and PLM-ICD are retrained on the same data splits as the agentic and cascade arms. All three use the pooled 5{,}000-note training set (SFT 2k + continuation 3k), the 500-note validation set for threshold tuning and early stopping, and the 1{,}000-note test set. A global decision threshold is selected by grid search over $[0.05, 0.925]$ in steps of 0.025, maximizing validation micro-F1; if no code exceeds the threshold for a given note, the argmax is taken. All experiments use a single NVIDIA L40S (46\,GB) GPU and random seed 3407.

\subsection{CAML (Mullenbach et al.\ 2018)}

Reimplementation in PyTorch 2.12.

\begin{table}[H]
\caption{CAML training configuration.}
\label{tab:caml}
\centering
\footnotesize
\begin{tabular}{@{}lll@{}}
\toprule
Parameter & Frequent (50) & All (15{,}761) \\
\midrule
Embedding dim       & 100 (word2vec, sg=1) & 100 (word2vec, sg=1) \\
Vocab size           & 21{,}615 & 21{,}597 \\
Vocab min count      & 3 & 3 \\
Max word tokens      & 4{,}000 & 4{,}000 \\
Conv kernel size     & 10 & 10 \\
Num filter maps      & 50 & 50 \\
Dropout              & 0.2 & 0.2 \\
Optimizer            & AdamW & AdamW \\
Learning rate        & 1e-3 & 1e-3 \\
Batch size           & 16 & 8 \\
Max epochs           & 50 & 50 \\
Early stop patience  & 5 & 5 \\
Gradient clip        & 1.0 & 1.0 \\
Loss                 & BCEWithLogitsLoss & BCEWithLogitsLoss \\
Best threshold       & 0.350 & 0.100 \\
Stopped at epoch     & 16 (best=11) & 6 (best=1) \\
\bottomrule
\end{tabular}
\end{table}

\subsection{LAAT (Vu et al.\ 2020)}

Reimplementation in PyTorch 2.12.

\begin{table}[H]
\caption{LAAT training configuration.}
\label{tab:laat}
\centering
\footnotesize
\begin{tabular}{@{}lll@{}}
\toprule
Parameter & Frequent (50) & All (15{,}761) \\
\midrule
Embedding dim       & 100 (word2vec, sg=1) & 100 (word2vec, sg=1) \\
Vocab size           & 21{,}615 & 21{,}597 \\
Vocab min count      & 3 & 3 \\
Max word tokens      & 4{,}000 & 4{,}000 \\
BiLSTM hidden        & 256/dir (512 total) & 256/dir (512 total) \\
Label attention $d_a$ & 256 & 256 \\
Dropout              & 0.3 & 0.3 \\
Optimizer            & AdamW & AdamW \\
Learning rate        & 1e-3 & 1e-3 \\
Batch size           & 16 & 8 \\
Max epochs           & 50 & 50 \\
Early stop patience  & 5 & 5 \\
Gradient clip        & 1.0 & 1.0 \\
Loss                 & BCEWithLogitsLoss & BCEWithLogitsLoss \\
Packed sequences     & Yes & Yes \\
Best threshold       & 0.350 & 0.300 \\
Stopped at epoch     & 13 (best=8) & 28 (best=23) \\
\bottomrule
\end{tabular}
\end{table}

\subsection{PLM-ICD (Huang et al.\ 2022)}
Reimplementation in PyTorch 2.12 and Transformers 5.x.

\begin{table}[H]
\caption{PLM-ICD training configuration.}
\label{tab:plmicd_hyper}
\centering
\footnotesize
\begin{tabular}{@{}lll@{}}
\toprule
Parameter & Frequent (50) & All (15{,}761) \\
\midrule
Encoder              & \multicolumn{2}{l}{BiomedBERT (microsoft/BiomedNLP-} \\
                     & \multicolumn{2}{l}{BiomedBERT-base-uncased-abstract-fulltext)} \\
Encoder hidden dim   & 768 & 768 \\
Max subword tokens   & 4{,}096 & 4{,}096 \\
Chunk size           & 128 & 128 \\
Label attention $d_a$ & 256 & 256 \\
Optimizer            & AdamW & AdamW \\
Learning rate        & 3e-5 & 3e-5 \\
Batch size           & 2 & 1 \\
Gradient accumulation & 4 & 8 \\
Effective batch size & 8 & 8 \\
LR schedule          & Linear warmup (10\%) & Same \\
                     & + linear decay & \\
Max epochs           & 15 & 15 \\
Early stop patience  & 4 & 4 \\
Gradient clip        & 1.0 & 1.0 \\
Loss                 & BCEWithLogitsLoss & BCEWithLogitsLoss \\
Best threshold       & 0.425 & 0.100 \\
Stopped at epoch     & 11 (best=7) & 11 (best=7) \\
\bottomrule
\end{tabular}
\end{table}